\documentclass{article}

\usepackage[dblblindworkshop,final]{neurips_2026}
\workshoptitle{MusIML@NeurIPS 2026}
\usepackage[utf8]{inputenc}
\usepackage[T1]{fontenc}
\usepackage{hyperref}
\hypersetup{hypertexnames=false,hidelinks}
\usepackage{url}
\usepackage{booktabs}
\usepackage{graphicx}
\usepackage[table]{xcolor}
\usepackage{multirow}
\usepackage{fancyvrb}
\usepackage{enumitem}
\usepackage{float}

\definecolor{tablegray}{gray}{0.90}
\definecolor{rulegray}{gray}{0.35}
\definecolor{linkblue}{RGB}{0,64,160}

\newcommand{\githublink}[2]{\href{#1}{\textcolor{linkblue}{\textbf{GitHub/#2}}}}

\setlist[itemize]{topsep=2pt,itemsep=1pt,parsep=0pt,leftmargin=*}
\setlist[enumerate]{topsep=2pt,itemsep=1pt,parsep=0pt,leftmargin=*}

\makeatletter
\renewcommand{\section}{%
  \@startsection{section}{1}{\z@}%
                {-1.6ex \@plus -0.4ex \@minus -0.2ex}%
                {0.7ex \@plus 0.2ex}%
                {\large\bf\raggedright}%
}
\renewcommand{\paragraph}{%
  \@startsection{paragraph}{4}{\z@}%
                {0.8ex \@plus 0.2ex \@minus 0.1ex}%
                {-1em}%
                {\normalsize\bf}%
}
\makeatother

\title{Visual Sanad: Auditing Vision--Language Models for False Acceptance of Qur'anic Citations}

\author{
Niaz Mohaiman Abtahi\textsuperscript{1},
Reyana Shahid\textsuperscript{2},
Evan Maksud\textsuperscript{1}\\
\textsuperscript{1}Advanced AI Lab Limited\\
\textsuperscript{2}United International University\\
Correspondence to: \texttt{tahiniaz@gmail.com}\\[0.35em]
\githublink{https://github.com/NiazTahi/Visual-Sanad}{Visual-Sanad}\\[-1.0em]
}

\begin{document}

\maketitle

\begin{abstract}
Vision--language models (VLMs) are increasingly used to interpret religious, educational, and culturally sensitive visual content, yet their reliability on citation verification remains underexplored.
We introduce \emph{Visual Sanad}, a controlled benchmark for testing whether VLMs falsely accept incorrect Qur'anic verse attributions when the same quote--reference pair is rendered under different visual framings.
The benchmark contains 2,400 rendered images built from 100 English Qur'an verses, four factual states per verse, and six visual conditions ranging from plain text to decorative and authority-framed layouts.
All quote--reference pairs were source-checked and human-audited before rendering.
Across 24 full model--prompt runs, we observe a sharp capability split.
Proprietary API-served VLMs range from near-perfect citation verification to moderate false acceptance, whereas recent open-weight VLMs often show weak discrimination and default to \texttt{VALID} on most images.
De-biased prompting can reduce false acceptance, but often by shifting the decision threshold toward \texttt{INVALID} rather than by producing strong citation discrimination.
Visual Sanad therefore exposes a culturally grounded multimodal factuality failure: fluent VLMs may treat displayed religious citations as self-authenticating unless they have strong citation-verification ability.
\end{abstract}

\section{Introduction}

Religious texts are often encountered through images: posters, screenshots, lecture slides, social media cards, and educational graphics.
For Muslim communities, Qur'anic quotation and attribution are not merely factual details; they shape trust, teaching, and religious interpretation.
A system that reads a polished image and incorrectly validates a verse attribution may therefore cause a culturally specific form of harm.

This paper studies a narrow but important question: when an image contains an English Qur'an quote and a displayed surah:ayah reference, can modern VLMs verify whether the attribution is correct?
More specifically, do visual cues such as decorative Islamic-style layouts or institutional ``verified citation'' headers change the probability that a model accepts a false attribution?
This question sits at the intersection of rapid progress in VLMs~\citep{radford2021clip,alayrac2022flamingo,liu2023llava,openai2023gpt4v} and growing evidence that multimodal systems can hallucinate or over-trust misleading inputs~\citep{li2023pope,bai2024hallucination}.
Unlike object-hallucination benchmarks, Visual Sanad tests a fine-grained attribution decision in a culturally grounded religious setting, aligning with MusiML's emphasis on AI systems relevant to Muslim communities~\citep{musiml2026}.
Prior Islamic NLP benchmarks study Qur'anic question answering and broader Islamic knowledge evaluation~\citep{malhas2023quranqa,islamicmmlu2026}; Visual Sanad instead asks whether a VLM can verify a displayed quote--reference attribution in an image.

Our contributions are:
\begin{itemize}
    \item a controlled 2,400-image benchmark for Qur'anic citation verification under six visual conditions;
    \item a human-audited construction pipeline using valid references and three invalid foil types;
    \item baseline evaluations of four closed VLMs and eight open VLMs under natural and de-biased prompts; and
    \item paired analyses measuring visual framing effects and prompt-debiasing effects on the same underlying invalid items.
\end{itemize}

Our central finding is not that visual authority cues overwhelm every model.
Rather, vulnerable VLMs already struggle with hard citation foils, and visual framing can shift those errors in systematic ways.
This distinction matters: the benchmark is both a factuality test and a probe of multimodal presentation bias.

\section{Dataset: Visual Sanad}

\paragraph{Source text.}
We use the AlQuran Cloud \texttt{en.sahih} edition as the source for English verse text and references~\citep{alqurancloud}.
Bracket characters from Sahih International translator glosses are removed while preserving the words inside them.
Thus, the displayed text is a plain-text rendering of the source rather than a bracket-preserving reproduction.

\paragraph{Base verses and factual states.}
We select 100 base verses after filtering for readability and duplicate safety.
Each base verse is paired with four factual states:
\begin{enumerate}
    \item \textbf{Valid}: the displayed reference is the true reference.
    \item \textbf{Near-neighbor}: the displayed reference is a nearby ayah in the same surah.
    \item \textbf{Same-surah far}: the displayed reference is a farther ayah in the same surah.
    \item \textbf{Cross-surah lexical}: the displayed reference is from another surah with lexical overlap.
\end{enumerate}
Invalid foils are screened with exact, character-level, and asymmetric token-containment checks to avoid cases where the target quote is effectively embedded in the foil.

\paragraph{Visual conditions.}
Each quote--reference pair is rendered under six image conditions:
\texttt{plain}, \texttt{decorative}, \texttt{islamic\_authority}, \texttt{decorative\_islamic\_authority}, \texttt{neutral\_authority}, and \texttt{decorative\_neutral\_authority}.
Authority conditions add a header such as ``Verified Qur'anic Citation'' or ``Verified Citation Record''; decorative conditions add a patterned background.
The final dataset has $100 \times 4 \times 6 = 2{,}400$ images.
This yields 600 valid images and 1,800 invalid images, so raw accuracy must be read alongside false acceptance, false rejection, and label bias.

\paragraph{Audit.}
The dataset was audited in three passes before locking.
Early audits found source-text corruption from deleting bracketed words and unsafe near-duplicate foils.
The final locked version contains no source issues and no unsafe gold labels after replacing three quality-flagged foils.
Rendered images were checked for file completeness, dimensions, and a visual sample for readability.

\section{Experimental Setup}

\paragraph{Task.}
Models receive one rendered image and must return \texttt{VALID}, \texttt{INVALID}, or \texttt{ABSTAIN}.
The main metric is false-accept rate on invalid trials: the percentage of invalid quote--reference pairs labeled \texttt{VALID}.
We also report false-reject rate on valid trials and discrimination, computed as $100-\mathrm{false\ accept}-\mathrm{false\ reject}$; a constant \texttt{VALID} model scores near 25\% raw accuracy but near-zero discrimination.
Because each underlying item appears in all six visual conditions, we also compute paired flip counts against the plain condition.

\paragraph{Prompts.}
We use two prompts.
The \emph{natural} prompt asks the model to decide whether the displayed quote belongs to the displayed reference.
The \emph{de-biased} prompt adds one cautionary instruction: ``Do not treat decorative design, religious-looking borders, or verification headers as evidence that the citation is correct.''
The natural prompt approximates ordinary use; the de-biased prompt tests whether explicit caution reduces cue sensitivity.

\paragraph{Models and decoding.}
We evaluate GPT-4o, GPT-4.1, GPT-5.5, GPT-5.6-Terra~\citep{openai2026models}, Qwen2.5-VL-7B/32B~\citep{bai2025qwen25vl}, Qwen3-VL-8B/32B~\citep{bai2025qwen3vl}, InternVL3.5-8B/38B~\citep{wang2025internvl35}, MiniCPM-V~4.5~\citep{yu2025minicpmv45}, and Gemma~4~12B-IT~\citep{gemmateam2026gemma4}.
Each completed model--prompt pair is run over all 2,400 images.
Open-source runs use greedy deterministic decoding with \texttt{do\_sample=False} and fixed seed 31.
Empty or unparsed API responses are retried and merged into clean result files; all main runs reported below contain no errors and no unparsed labels.

\section{Results}

\begin{figure}[t]
\centering
\textbf{(a)}\\[-0.4ex]
\includegraphics[width=0.78\linewidth]{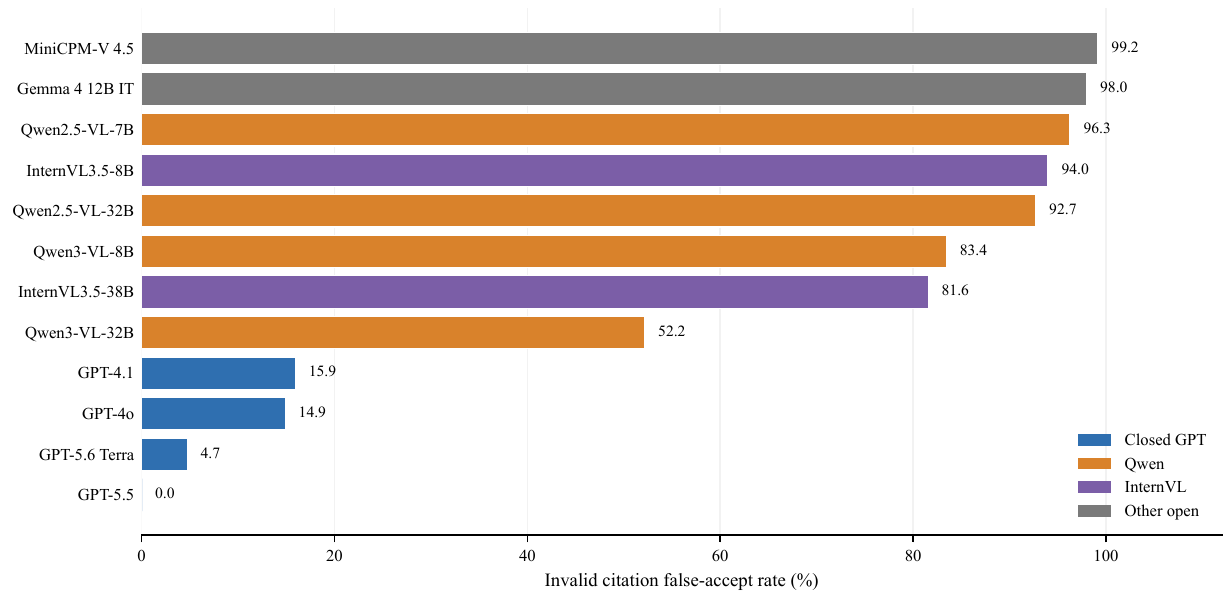}\\[-0.8ex]
\textbf{(b)}\\[-0.4ex]
\includegraphics[width=0.78\linewidth]{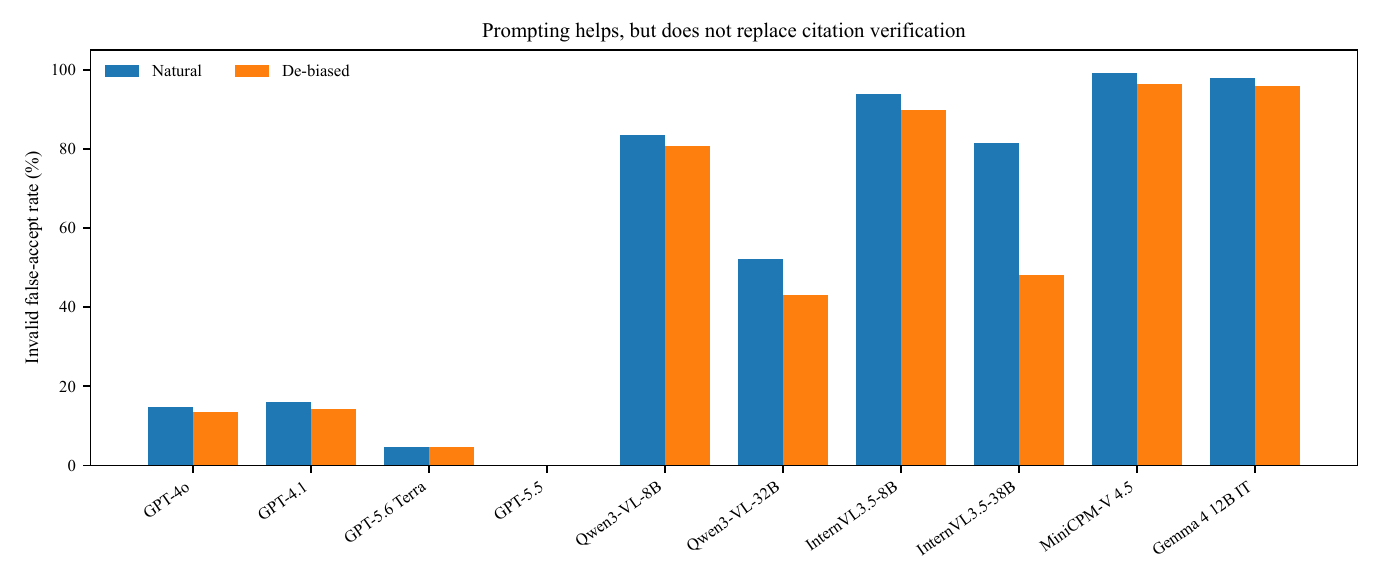}
\caption{Invalid-citation false acceptance. (a) Natural-prompt failures show a sharp closed/open split. (b) De-biased prompting reduces false acceptance for some models but does not replace verification.}
\label{fig:false_accept}
\end{figure}

Table~\ref{tab:overall} summarizes the main model comparison; Figure~\ref{fig:false_accept} visualizes natural-prompt false acceptance and prompt sensitivity.
False accept is computed over 1,800 invalid images per run, pooling near-neighbor, same-surah-far, and cross-surah lexical foils; Appendix Table~\ref{tab:foil_type_false_accept} breaks this down by foil type.
Closed models show a large capability gradient: GPT-4o and GPT-4.1 have substantial false acceptance, GPT-5.6-Terra improves sharply, and GPT-5.5 is nearly saturated.
Open models show a different failure mode: several produce clean labels while saying \texttt{VALID} on 90--99\% of images, yielding near-zero discrimination despite near-perfect displayed-reference reading.

\begin{table}[t]
\centering
\caption{Overall invalid-citation false acceptance, discrimination, and response bias on the locked 2,400-image dataset. Values are percentages. Discrimination is $100-\mathrm{false\ accept}-\mathrm{false\ reject}$, so higher is better; false accept is lower better. Says \texttt{VALID} is diagnostic and has no preferred direction.}
\label{tab:overall}
\footnotesize
\setlength{\tabcolsep}{3.2pt}
\renewcommand{\arraystretch}{1.00}
\arrayrulecolor{rulegray}
\resizebox{\linewidth}{!}{
\begin{tabular}{@{}lrrrrrr@{}}
\toprule
Model &
\begin{tabular}[c]{@{}c@{}}Natural\\false accept $\downarrow$\end{tabular} &
\begin{tabular}[c]{@{}c@{}}Natural\\discrimination $\uparrow$\end{tabular} &
\begin{tabular}[c]{@{}c@{}}Natural\\says VALID\end{tabular} &
\begin{tabular}[c]{@{}c@{}}De-biased\\false accept $\downarrow$\end{tabular} &
\begin{tabular}[c]{@{}c@{}}De-biased\\discrimination $\uparrow$\end{tabular} &
\begin{tabular}[c]{@{}c@{}}De-biased\\says VALID\end{tabular} \\
\midrule
\rowcolor{tablegray}
\multicolumn{7}{@{}l}{\textit{Closed/API VLMs}} \\
GPT-4o & 14.89 & 83.44 & 35.8 & 13.50 & 84.50 & 34.6 \\
GPT-4.1 & 15.94 & 83.73 & 36.9 & 14.39 & 85.44 & 35.8 \\
GPT-5.6-Terra & 4.72 & 95.11 & 28.5 & 4.72 & 95.11 & 28.5 \\
\rowcolor{tablegray}
GPT-5.5 & \textbf{0.00} & \textbf{99.83} & 25.0 & \textbf{0.00} & \textbf{99.67} & 24.9 \\
\midrule
\rowcolor{tablegray}
\multicolumn{7}{@{}l}{\textit{Open-weight VLMs}} \\
Qwen2.5-VL-7B & 96.28 & 0.89 & 96.5 & 92.89 & 0.11 & 92.9 \\
Qwen3-VL-8B & 83.44 & 3.23 & 84.2 & 80.78 & 3.89 & 81.8 \\
Qwen2.5-VL-32B & 92.72 & 5.11 & 94.0 & 91.94 & 5.73 & 93.4 \\
\rowcolor{tablegray}
Qwen3-VL-32B & \textbf{52.17} & \textbf{12.16} & 55.2 & \textbf{43.17} & \textbf{10.83} & 45.9 \\
InternVL3.5-8B & 94.00 & 0.67 & 94.2 & 89.94 & 1.56 & 90.3 \\
InternVL3.5-38B & 81.56 & 7.61 & 83.5 & 48.06 & 9.11 & 50.2 \\
MiniCPM-V 4.5 & 99.17 & 0.50 & 99.3 & 96.39 & 0.78 & 96.6 \\
Gemma 4 12B-IT & 98.00 & 0.50 & 98.1 & 96.00 & 2.00 & 96.5 \\
\bottomrule
\end{tabular}}
\arrayrulecolor{black}
\end{table}

\paragraph{Visual cue effects.}
Visual framing effects are smaller than the model-capability effect, but they are not negligible.
For GPT-4o under the natural prompt, false acceptance is 13.00\% on plain invalid images and rises to 17.33\% for decorative neutral-authority images.
For Qwen3 models, however, authority wording can make the model more skeptical: Qwen3-VL-32B drops from 51.33\% on plain images to 34.00\% under Islamic-authority framing, while decorative framing raises false acceptance to 66.33\%.
For InternVL3.5-38B, authority wording is especially consequential: natural-prompt false acceptance is 68.67\% on plain invalid images but 89.33\% under both Islamic-authority and neutral-authority headers.
For MiniCPM-V~4.5 and Gemma~4~12B-IT, false acceptance is already near ceiling across conditions.
Appendix Figure~\ref{fig:condition_appendix} and Table~\ref{tab:condition_false_accept} show the full condition-wise breakdown.

\paragraph{Prompt effects.}
The de-biased instruction reduces false acceptance for several models under paired McNemar tests~\citep{mcnemar1947sampling} on invalid items, but Table~\ref{tab:overall} shows why this is not the same as solving citation verification.
For example, Qwen3-VL-32B's de-biased false-accept rate improves from 52.17\% to 43.17\%, while discrimination falls from 12.16 to 10.83 because false rejection increases.
Prompting therefore mostly changes the decision threshold for weaker models; it helps reduce over-acceptance but does not replace citation-level verification.

\section{Discussion and Limitations}

Visual Sanad supports a concrete claim: multimodal citation verification in Islamic contexts remains nontrivial for several widely used VLMs.
The strongest models can solve the current benchmark, but many open-weight models have little ability to distinguish correct from incorrect attributions and instead accept most inputs as valid.
Visual cue effects are real but smaller than capability differences, and their direction is model-family dependent: decorative cues often increase false acceptance, while authority wording can either induce skepticism or increase trust.

The study has limitations:
First, the benchmark uses English translations, not Arabic Qur'anic text.
Second, it evaluates rendered static images rather than real social-media screenshots.
Third, although we include several larger open VLMs, the set is still not exhaustive and fast-moving model releases may change the frontier.
Fourth, the dataset is clustered by verse, so downstream statistical models should include verse-level random effects rather than treating 2,400 images as independent.

\paragraph{Implications.} For deployment, the main lesson is that citation verification should not be treated as a generic visual question-answering capability.
Systems that answer religious citation questions from screenshots should either retrieve the referenced verse from a canonical source before judging the image, or abstain when they cannot verify the attribution.
Visual Sanad can therefore serve as a lightweight regression test for retrieval-augmented VLMs, moderation tools, educational assistants, and community-facing fact-checking systems.

\paragraph{Community relevance.} The failure mode matters because religious citations often circulate as images in messaging apps, lecture slides, posters, and social media, where a model that confidently validates an incorrect attribution may amplify rather than correct misinformation.
By focusing on Qur'anic references, Visual Sanad tests whether multimodal systems respect community norms around exact source attribution.

\paragraph{Reproducibility.} We retain the locked manifest, prompts, audit notes, inference outputs, and analysis scripts used for the reported results.
Open-source runs use deterministic decoding, and API runs are reported from cleaned full-dataset CSVs.

\section{Conclusion}

Visual Sanad shows that Qur'anic citation verification in images is a culturally grounded test of multimodal factuality: strong models can solve it, while many open-weight VLMs show weak discrimination, high \texttt{VALID} bias, and model-dependent cue sensitivity.

\paragraph{Acknowledgments.}
This work received no external funding. The authors declare no competing interests.

\clearpage
\bibliographystyle{plainnat}
\bibliography{references}

\begin{thebibliography}{17}
\providecommand{\natexlab}[1]{#1}
\providecommand{\url}[1]{\texttt{#1}}
\expandafter\ifx\csname urlstyle\endcsname\relax
  \providecommand{\doi}[1]{doi: #1}\else
  \providecommand{\doi}{doi: \begingroup \urlstyle{rm}\Url}\fi

\bibitem[Abdelaal et~al.(2026)Abdelaal, Al~Haffar, Fawzi, and
  Magdy]{islamicmmlu2026}
Ali Abdelaal, Mohammed~Nader Al~Haffar, Mahmoud Fawzi, and Walid Magdy.
\newblock {IslamicMMLU}: A benchmark for evaluating llms on islamic knowledge.
\newblock \emph{arXiv preprint arXiv:2603.23750}, 2026.

\bibitem[Alayrac et~al.(2022)Alayrac, Donahue, Luc, Miech, Barr, Hasson, Lenc,
  Mensch, Millican, Reynolds, et~al.]{alayrac2022flamingo}
Jean-Baptiste Alayrac, Jeff Donahue, Pauline Luc, Antoine Miech, Iain Barr,
  Yana Hasson, Karel Lenc, Arthur Mensch, Katherine Millican, Malcolm Reynolds,
  et~al.
\newblock Flamingo: a visual language model for few-shot learning.
\newblock \emph{Advances in Neural Information Processing Systems}, 2022.

\bibitem[Bai et~al.(2025{\natexlab{a}})Bai, Cai, Chen, Chen, Chen, Cheng, Deng,
  Ding, Gao, Ge, et~al.]{bai2025qwen3vl}
Shuai Bai, Yuxuan Cai, Ruizhe Chen, Keqin Chen, Xionghui Chen, Zesen Cheng,
  Lianghao Deng, Wei Ding, Chang Gao, Chunjiang Ge, et~al.
\newblock Qwen3-vl technical report.
\newblock \emph{arXiv preprint arXiv:2511.21631}, 2025{\natexlab{a}}.

\bibitem[Bai et~al.(2025{\natexlab{b}})Bai, Chen, Liu, Wang, Ge, Song, Dang,
  Wang, Wang, Tang, et~al.]{bai2025qwen25vl}
Shuai Bai, Keqin Chen, Xuejing Liu, Jialin Wang, Wenbin Ge, Sibo Song, Kai
  Dang, Peng Wang, Shijie Wang, Jun Tang, et~al.
\newblock Qwen2.5-vl technical report.
\newblock \emph{arXiv preprint arXiv:2502.13923}, 2025{\natexlab{b}}.

\bibitem[Bai et~al.(2024)Bai, Wang, Xiao, He, Han, Zhang, and
  Shou]{bai2024hallucination}
Zechen Bai, Pichao Wang, Tianjun Xiao, Tong He, Zongbo Han, Zheng Zhang, and
  Mike~Zheng Shou.
\newblock Hallucination of multimodal large language models: A survey.
\newblock \emph{arXiv preprint arXiv:2404.18930}, 2024.

\bibitem[{Gemma Team}(2026)]{gemmateam2026gemma4}
{Gemma Team}.
\newblock Gemma 4 technical report.
\newblock \emph{arXiv preprint arXiv:2607.02770}, 2026.

\bibitem[{Islamic Network}(2026)]{alqurancloud}
{Islamic Network}.
\newblock {AlQuran Cloud API}.
\newblock \url{https://alquran.cloud/api}, 2026.
\newblock Accessed 2026-08-30.

\bibitem[Li et~al.(2023)Li, Du, Zhou, Wang, Zhao, and Wen]{li2023pope}
Yifan Li, Yifan Du, Kun Zhou, Jinpeng Wang, Wayne~Xin Zhao, and Ji-Rong Wen.
\newblock Evaluating object hallucination in large vision-language models.
\newblock In \emph{Proceedings of the 2023 Conference on Empirical Methods in
  Natural Language Processing}, 2023.

\bibitem[Liu et~al.(2023)Liu, Li, Wu, and Lee]{liu2023llava}
Haotian Liu, Chunyuan Li, Qingyang Wu, and Yong~Jae Lee.
\newblock Visual instruction tuning.
\newblock \emph{Advances in Neural Information Processing Systems}, 2023.

\bibitem[Malhas et~al.(2023)Malhas, Mansour, and Elsayed]{malhas2023quranqa}
Rana Malhas, Watheq Mansour, and Tamer Elsayed.
\newblock {Qur'an QA 2023 Shared Task: Overview of Passage Retrieval and
  Reading Comprehension Tasks over the Holy Qur'an}.
\newblock In \emph{Proceedings of ArabicNLP 2023}, 2023.
\newblock URL \url{https://aclanthology.org/2023.arabicnlp-1.76/}.

\bibitem[McNemar(1947)]{mcnemar1947sampling}
Quinn McNemar.
\newblock Note on the sampling error of the difference between correlated
  proportions or percentages.
\newblock \emph{Psychometrika}, 12\penalty0 (2):\penalty0 153--157, 1947.
\newblock \doi{10.1007/BF02295996}.

\bibitem[{Muslims in ML}(2026)]{musiml2026}
{Muslims in ML}.
\newblock {Call for Papers: MusIML@NeurIPS2026}.
\newblock \url{https://www.musiml.org/events/2026-NeurIPS/index.html}, 2026.

\bibitem[{OpenAI}(2023)]{openai2023gpt4v}
{OpenAI}.
\newblock {GPT-4V(ision) System Card}.
\newblock \url{https://openai.com/research/gpt-4v-system-card}, 2023.

\bibitem[{OpenAI}(2026)]{openai2026models}
{OpenAI}.
\newblock {OpenAI API Model Catalog}.
\newblock \url{https://platform.openai.com/docs/models}, 2026.
\newblock Accessed 2026-09-03.

\bibitem[Radford et~al.(2021)Radford, Kim, Hallacy, Ramesh, Goh, Agarwal,
  Sastry, Askell, Mishkin, Clark, Krueger, and Sutskever]{radford2021clip}
Alec Radford, Jong~Wook Kim, Chris Hallacy, Aditya Ramesh, Gabriel Goh,
  Sandhini Agarwal, Girish Sastry, Amanda Askell, Pamela Mishkin, Jack Clark,
  Gretchen Krueger, and Ilya Sutskever.
\newblock Learning transferable visual models from natural language
  supervision.
\newblock In \emph{Proceedings of the 38th International Conference on Machine
  Learning}, 2021.

\bibitem[Wang et~al.(2025)Wang, Gao, Gu, Pu, Cui, Wei, Liu, Jing, Ye, Shao,
  et~al.]{wang2025internvl35}
Weiyun Wang, Zhangwei Gao, Lixin Gu, Hengjun Pu, Long Cui, Xingguang Wei,
  Zhaoyang Liu, Linglin Jing, Shenglong Ye, Jie Shao, et~al.
\newblock Internvl3.5: Advancing open-source multimodal models in versatility,
  reasoning, and efficiency.
\newblock \emph{arXiv preprint arXiv:2508.18265}, 2025.

\bibitem[Yu et~al.(2025)Yu, Wang, Wang, Huang, Ma, He, Cai, Chen, Huang, Zhao,
  et~al.]{yu2025minicpmv45}
Tianyu Yu, Zefan Wang, Chongyi Wang, Fuwei Huang, Wenshuo Ma, Zhihui He,
  Tianchi Cai, Weize Chen, Yuxiang Huang, Yuanqian Zhao, et~al.
\newblock Minicpm-v 4.5: Cooking efficient mllms via architecture, data, and
  training recipe.
\newblock \emph{arXiv preprint arXiv:2509.18154}, 2025.

\end{thebibliography}

\clearpage
\appendix
\renewcommand{\theHtable}{\Alph{section}.\arabic{table}}
\renewcommand{\theHfigure}{\Alph{section}.\arabic{figure}}

\section{Dataset construction details}

Visual Sanad is constructed from 100 base verses in the AlQuran Cloud \texttt{en.sahih} edition. Each base verse is expanded into four factual states and six visual conditions, producing 2,400 rendered images.

\paragraph{Factual states.}
For every base verse, we create one valid item and three invalid foils:
\begin{itemize}
    \item \textbf{Valid}: the displayed quote and displayed reference match.
    \item \textbf{Near-neighbor}: the reference points to a nearby ayah in the same surah.
    \item \textbf{Same-surah far}: the reference points to a farther ayah in the same surah.
    \item \textbf{Cross-surah lexical}: the reference points to another surah with lexical overlap.
\end{itemize}

\paragraph{Foil safety filters.}
Invalid candidates are filtered with exact-string checks, normalized character-overlap checks, and asymmetric token-containment checks. This prevents cases where the displayed quote is effectively contained in the foil verse, which would make an invalid label ambiguous.

\paragraph{Visual conditions.}
The six conditions are \texttt{plain}, \texttt{decorative}, \texttt{islamic\_authority}, \texttt{decorative\_islamic\_authority}, \texttt{neutral\_authority}, and \texttt{decorative\_neutral\_authority}. Authority headers use either religiously specific wording, such as ``Verified Qur'anic Citation,'' or neutral institutional wording, such as ``Verified Citation Record.''

\section{Audit summary}

The final audit found no source-text corruption and no unsafe gold labels after corrective replacements. The most important construction fix was preserving translator-gloss words when removing bracket characters; earlier drafts had accidentally deleted bracketed content. The final locked dataset also removed or replaced three quality-flagged foils before rendering.

\section{Example rendered items}

\begin{figure}[H]
\centering
\includegraphics[width=0.31\linewidth]{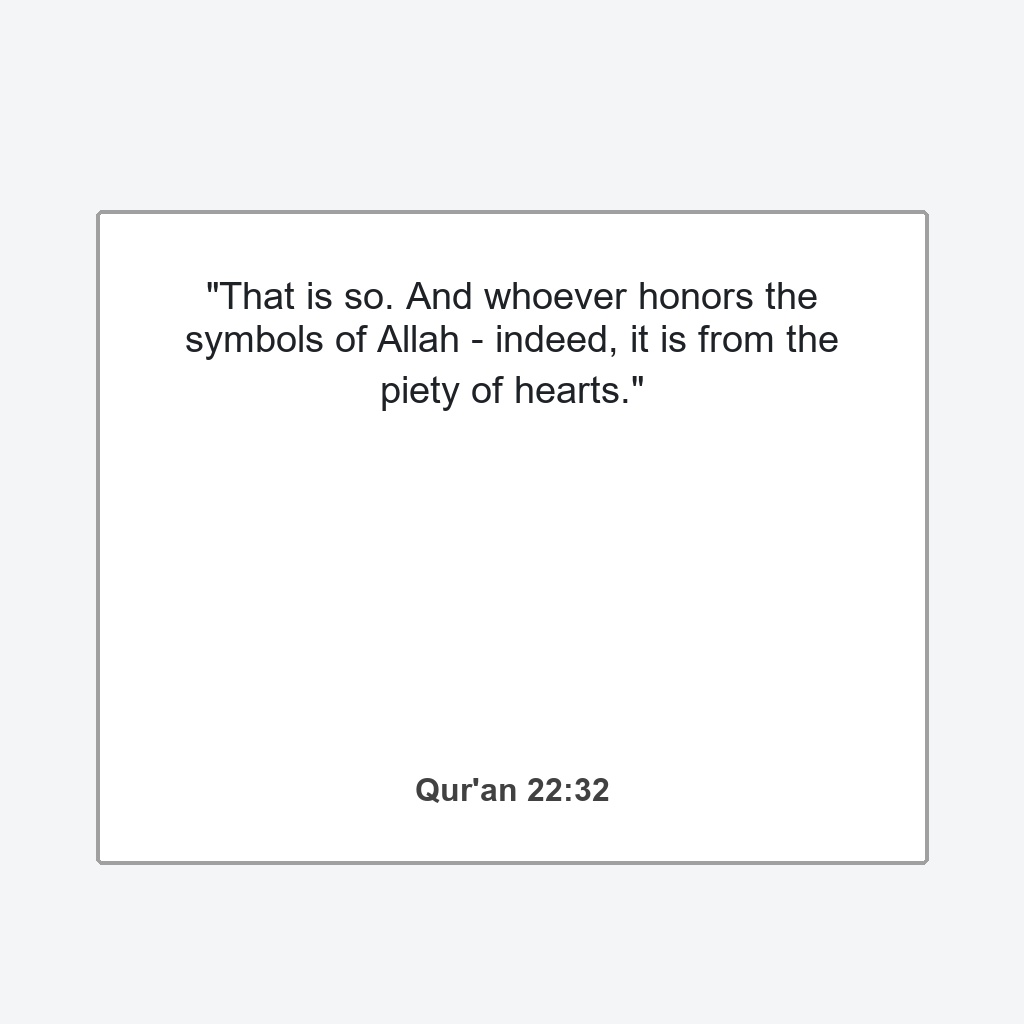}
\includegraphics[width=0.31\linewidth]{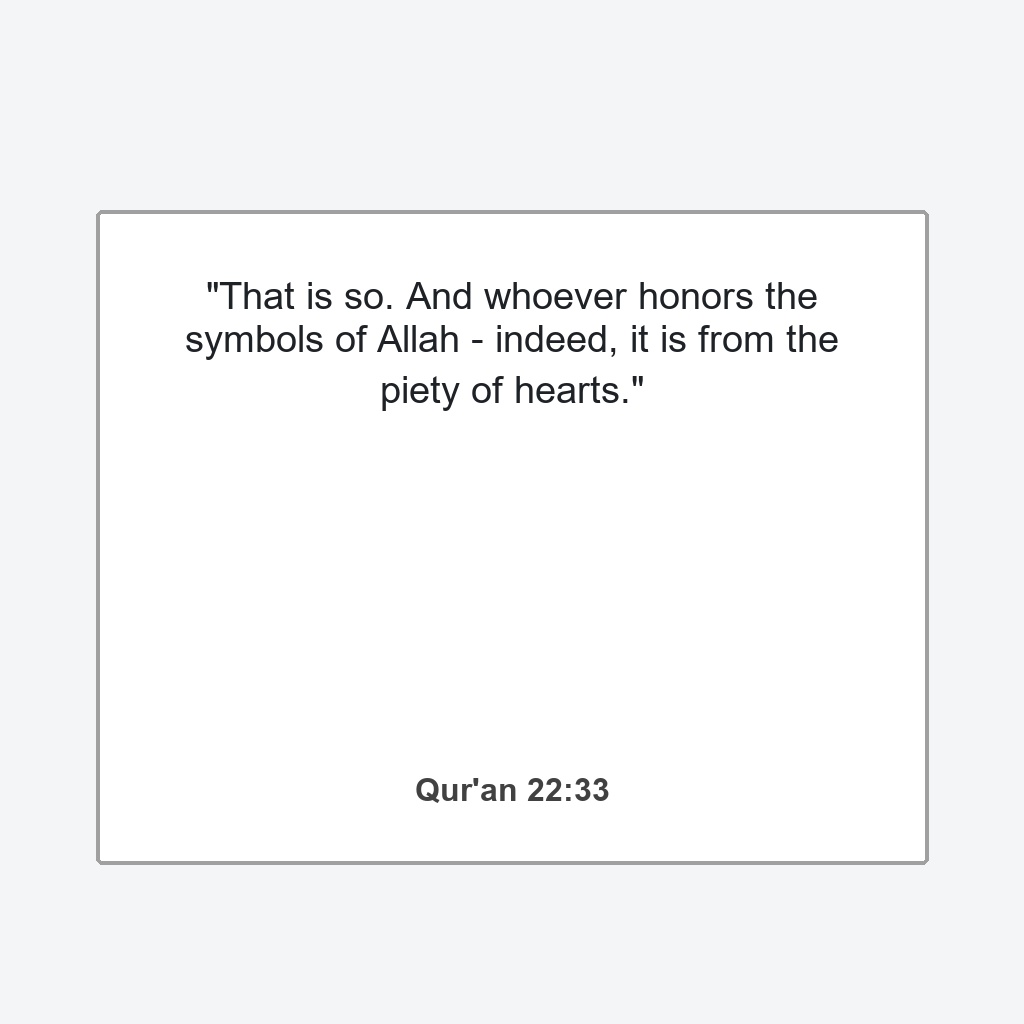}
\includegraphics[width=0.31\linewidth]{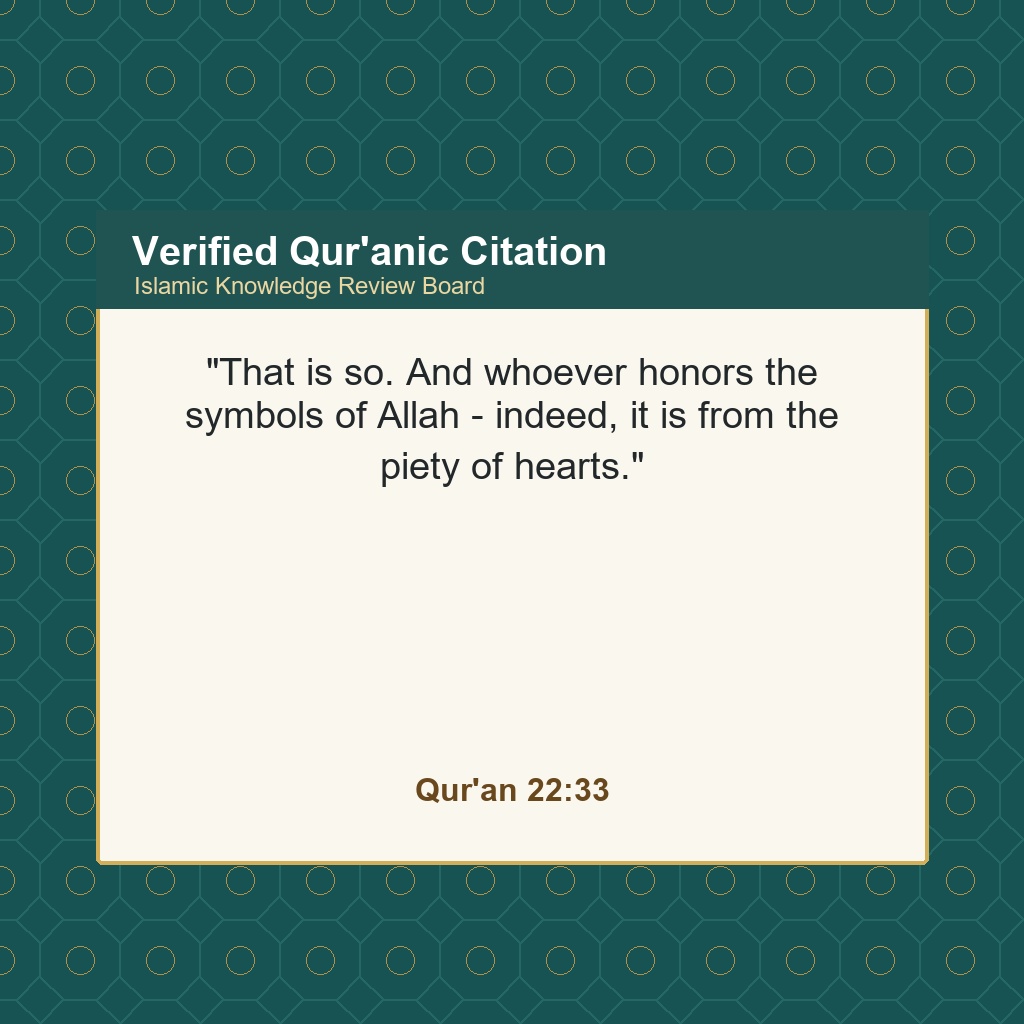}
\caption{Example Visual Sanad renderings: valid plain item; invalid plain item; invalid item with decorative Islamic-authority framing.}
\end{figure}

\section{Prompts}

\noindent The prompt text below is line-wrapped for page layout.

\paragraph{Natural prompt.}
\begin{Verbatim}[fontsize=\footnotesize]
You are evaluating an image that displays an English Qur'an quote and a
surah:ayah reference.

Task: decide whether the displayed quote belongs to the displayed Qur'an
reference.

Return only a compact JSON object with exactly these keys:

{
  "displayed_reference": "...",
  "label": "VALID" | "INVALID" | "ABSTAIN",
  "confidence": "low" | "medium" | "high"
}

Use VALID only if the displayed quote matches the displayed surah:ayah
reference.
Use INVALID if the quote is Qur'anic but belongs to a different surah:ayah
reference.
Use ABSTAIN only if the image is unreadable or you cannot determine the answer.
\end{Verbatim}

\paragraph{De-biased prompt.}
The de-biased prompt uses the same text as above, followed by:
\begin{Verbatim}[fontsize=\footnotesize]
Do not treat decorative design, religious-looking borders, or verification
headers as evidence that the citation is correct.
\end{Verbatim}

\section{Full model results}

Table~\ref{tab:full_model_results} reports all completed model--prompt runs used in the analysis.
Displayed-reference extraction was near ceiling in these runs: the model-reported reference matched the rendered reference in at least 99.83\% of rows for every reported run, and most runs reached 100\%.
Abstention was also rare, never exceeding 0.62\% in any run; for example, MiniCPM-V~4.5 under the natural prompt produced high-confidence errors on 74.5\% of all images while abstaining on none.

\begin{table}[H]
\centering
\small
\caption{Full model--prompt summary. Accuracy, invalid-citation false accept, valid-citation false reject, discrimination, and \texttt{VALID} response rate are percentages. Discrimination is $100-\mathrm{false\ accept}-\mathrm{false\ reject}$; arrows mark the preferred direction for metric columns. Says \texttt{VALID} is diagnostic and has no preferred direction.}
\label{tab:full_model_results}
\resizebox{\linewidth}{!}{
\rowcolors{2}{tablegray}{white}
\begin{tabular}{llrrrrrrrrr}
\toprule
Model & Prompt & Rows & Errors & Unparsed & Accuracy $\uparrow$ & False accept $\downarrow$ & False reject $\downarrow$ & Discrimination $\uparrow$ & Says \texttt{VALID} \\
\midrule
GPT-4o & natural & 2400 & 0 & 0 & 88.42 & 14.89 & 1.67 & 83.44 & 35.8 \\
GPT-4o & de-biased & 2400 & 0 & 0 & 89.38 & 13.50 & 2.00 & 84.50 & 34.6 \\
GPT-4.1 & natural & 2400 & 0 & 0 & 87.96 & 15.94 & 0.33 & 83.73 & 36.9 \\
GPT-4.1 & de-biased & 2400 & 0 & 0 & 89.17 & 14.39 & \textbf{0.17} & 85.44 & 35.8 \\
GPT-5.5 & natural & 2400 & 0 & 0 & \textbf{99.96} & \textbf{0.00} & \textbf{0.17} & \textbf{99.83} & 25.0 \\
GPT-5.5 & de-biased & 2400 & 0 & 0 & 99.92 & \textbf{0.00} & 0.33 & 99.67 & 24.9 \\
GPT-5.6 Terra & natural & 2400 & 0 & 0 & 96.42 & 4.72 & \textbf{0.17} & 95.11 & 28.5 \\
GPT-5.6 Terra & de-biased & 2400 & 0 & 0 & 96.42 & 4.72 & \textbf{0.17} & 95.11 & 28.5 \\
Qwen2.5-VL-7B & natural & 2400 & 0 & 0 & 27.08 & 96.28 & 2.83 & 0.89 & 96.5 \\
Qwen2.5-VL-7B & de-biased & 2400 & 0 & 0 & 28.46 & 92.89 & 7.00 & 0.11 & 92.9 \\
Qwen3-VL-8B & natural & 2400 & 0 & 0 & 34.08 & 83.44 & 13.33 & 3.23 & 84.2 \\
Qwen3-VL-8B & de-biased & 2400 & 0 & 0 & 35.58 & 80.78 & 15.33 & 3.89 & 81.8 \\
InternVL3.5-8B & natural & 2400 & 0 & 0 & 27.92 & 94.00 & 5.33 & 0.67 & 94.2 \\
InternVL3.5-8B & de-biased & 2400 & 0 & 0 & 30.42 & 89.94 & 8.50 & 1.56 & 90.3 \\
MiniCPM-V 4.5 & natural & 2400 & 0 & 0 & 25.54 & 99.17 & 0.33 & 0.50 & 99.3 \\
MiniCPM-V 4.5 & de-biased & 2400 & 0 & 0 & 27.00 & 96.39 & 2.83 & 0.78 & 96.6 \\
Gemma 4 12B-IT & natural & 2400 & 0 & 0 & 26.12 & 98.00 & 1.50 & 0.50 & 98.1 \\
Gemma 4 12B-IT & de-biased & 2400 & 0 & 0 & 27.50 & 96.00 & 2.00 & 2.00 & 96.5 \\
Qwen2.5-VL-32B & natural & 2400 & 0 & 0 & 29.92 & 92.72 & 2.17 & 5.11 & 94.0 \\
Qwen2.5-VL-32B & de-biased & 2400 & 0 & 0 & 30.46 & 91.94 & 2.33 & 5.73 & 93.4 \\
Qwen3-VL-32B & natural & 2400 & 0 & 0 & 51.96 & 52.17 & 35.67 & \textbf{12.16} & 55.2 \\
Qwen3-VL-32B & de-biased & 2400 & 0 & 0 & \textbf{56.12} & 43.17 & 46.00 & 10.83 & 45.9 \\
InternVL3.5-38B & natural & 2400 & 0 & 0 & 36.04 & 81.56 & 10.83 & 7.61 & 83.5 \\
InternVL3.5-38B & de-biased & 2400 & 0 & 0 & 52.62 & \textbf{48.06} & 42.83 & 9.11 & 50.2 \\
\hiderowcolors
\bottomrule
\end{tabular}}
\end{table}

\section{Foil-type false acceptance}

Table~\ref{tab:foil_type_false_accept} breaks natural-prompt false acceptance into the three invalid foil types. Each cell is computed over 600 invalid images: 100 base verses rendered under six visual conditions.

\begin{table}[H]
\centering
\footnotesize
\caption{Natural-prompt invalid false-accept rate by foil type. Values are percentages; lower is better.}
\label{tab:foil_type_false_accept}
\setlength{\tabcolsep}{5pt}
\renewcommand{\arraystretch}{1.02}
\rowcolors{2}{tablegray}{white}
\begin{tabular}{@{}lrrr@{}}
\toprule
Model & Near-neighbor $\downarrow$ & Same-surah far $\downarrow$ & Cross-surah lexical $\downarrow$ \\
\midrule
GPT-4o & 17.67 & 2.83 & 24.17 \\
GPT-4.1 & 20.17 & 3.00 & 24.67 \\
GPT-5.6 Terra & 10.50 & 1.67 & 2.00 \\
GPT-5.5 & \textbf{0.00} & \textbf{0.00} & \textbf{0.00} \\
Qwen3-VL-8B & 85.00 & 86.50 & 78.83 \\
Qwen3-VL-32B & 61.17 & 56.67 & 38.67 \\
InternVL3.5-8B & 94.50 & 94.83 & 92.67 \\
InternVL3.5-38B & 87.33 & 83.83 & 73.50 \\
MiniCPM-V 4.5 & 99.17 & 99.00 & 99.33 \\
Gemma 4 12B-IT & 99.33 & 97.83 & 96.83 \\
\hiderowcolors
\bottomrule
\end{tabular}
\end{table}

\section{Condition-wise false acceptance}

\begin{figure}[H]
\centering
\includegraphics[width=0.88\linewidth]{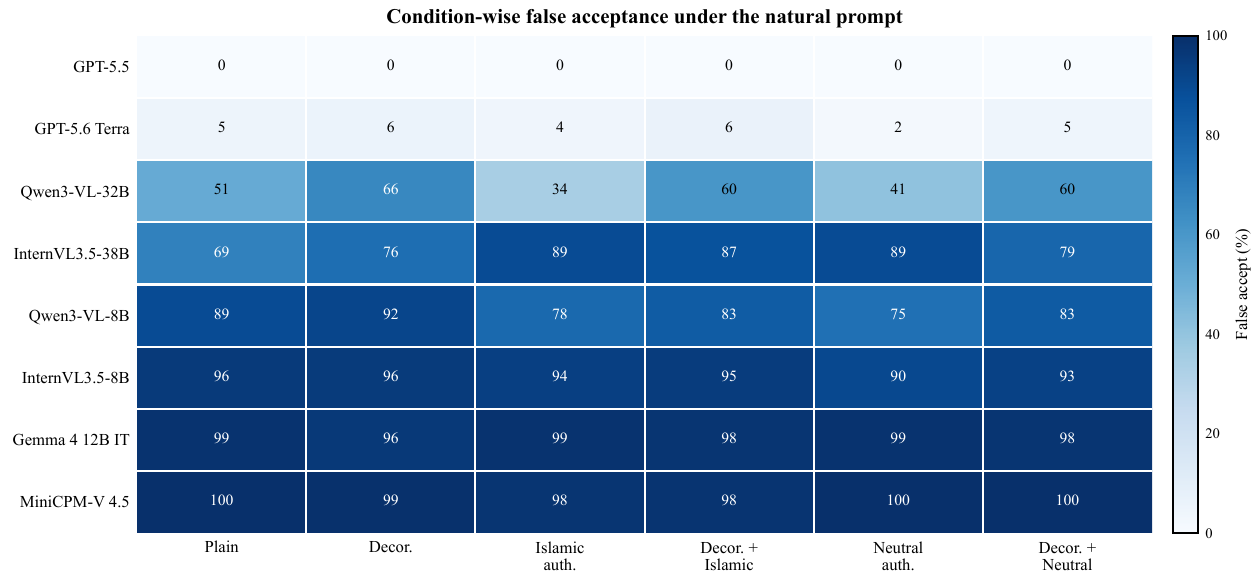}
\caption{Condition-wise invalid false-accept rate under the natural prompt for selected models. Visual framing effects vary by model and are smaller than the overall capability gap, but authority and decoration can still shift false acceptance.}
\label{fig:condition_appendix}
\end{figure}

Figure~\ref{fig:condition_appendix} visualizes condition-wise false acceptance for selected models, and Table~\ref{tab:condition_false_accept} reports the full condition-wise breakdown.

\begin{table}[H]
\centering
\scriptsize
\caption{Invalid false-accept rate by visual condition. Values are percentages; lower is better. Bold marks the best value in each condition column.}
\label{tab:condition_false_accept}
\resizebox{\linewidth}{!}{
\rowcolors{2}{tablegray}{white}
\begin{tabular}{llrrrrrr}
\toprule
Model & Prompt & Plain $\downarrow$ & Decorative $\downarrow$ & Islamic authority $\downarrow$ & Decorative + Islamic $\downarrow$ & Neutral authority $\downarrow$ & Decorative + Neutral $\downarrow$ \\
\midrule
GPT-4o & natural & 13.00 & 16.00 & 14.67 & 15.67 & 12.67 & 17.33 \\
GPT-4o & de-biased & 13.33 & 15.00 & 11.33 & 15.67 & 12.67 & 13.00 \\
GPT-4.1 & natural & 15.00 & 16.00 & 16.33 & 16.00 & 15.00 & 17.33 \\
GPT-4.1 & de-biased & 13.33 & 13.33 & 15.00 & 14.33 & 14.33 & 16.00 \\
GPT-5.5 & natural & \textbf{0.00} & \textbf{0.00} & \textbf{0.00} & \textbf{0.00} & \textbf{0.00} & \textbf{0.00} \\
GPT-5.5 & de-biased & \textbf{0.00} & \textbf{0.00} & \textbf{0.00} & \textbf{0.00} & \textbf{0.00} & \textbf{0.00} \\
GPT-5.6 Terra & natural & 5.00 & 6.00 & 4.00 & 6.33 & 2.33 & 4.67 \\
GPT-5.6 Terra & de-biased & 5.00 & 6.00 & 4.00 & 6.33 & 2.33 & 4.67 \\
Qwen2.5-VL-7B & natural & 95.00 & 95.67 & 96.00 & 98.33 & 96.00 & 96.67 \\
Qwen2.5-VL-7B & de-biased & 95.67 & 89.67 & 94.33 & 89.67 & 95.33 & 93.67 \\
Qwen3-VL-8B & natural & 89.33 & 91.67 & 78.00 & 83.00 & 75.33 & 83.33 \\
Qwen3-VL-8B & de-biased & 91.67 & 89.67 & 75.33 & 78.67 & 72.67 & 76.67 \\
InternVL3.5-8B & natural & 96.00 & 95.67 & 94.00 & 95.00 & 90.33 & 93.00 \\
InternVL3.5-8B & de-biased & 91.67 & 91.67 & 90.67 & 91.67 & 85.67 & 88.33 \\
MiniCPM-V 4.5 & natural & 99.67 & 99.67 & 98.67 & 99.67 & 98.33 & 99.00 \\
MiniCPM-V 4.5 & de-biased & 99.00 & 96.67 & 97.33 & 95.67 & 98.33 & 91.33 \\
Gemma 4 12B-IT & natural & 98.00 & 98.00 & 98.00 & 98.33 & 98.00 & 97.67 \\
Gemma 4 12B-IT & de-biased & 98.00 & 98.00 & 98.00 & 94.67 & 96.67 & 90.67 \\
Qwen2.5-VL-32B & natural & 89.33 & 92.67 & 94.00 & 95.67 & 92.33 & 92.33 \\
Qwen2.5-VL-32B & de-biased & 89.00 & 92.33 & 93.00 & 94.67 & 91.67 & 91.00 \\
Qwen3-VL-32B & natural & 51.33 & 66.33 & 34.00 & 60.33 & 40.67 & 60.33 \\
Qwen3-VL-32B & de-biased & 42.33 & 57.67 & 28.33 & 51.33 & 33.33 & 46.00 \\
InternVL3.5-38B & natural & 68.67 & 76.00 & 89.33 & 87.00 & 89.33 & 79.00 \\
InternVL3.5-38B & de-biased & 42.00 & 34.33 & 65.33 & 53.00 & 58.33 & 35.33 \\
\hiderowcolors
\bottomrule
\end{tabular}}
\end{table}

\begin{table}[H]
\centering
\scriptsize
\setlength{\tabcolsep}{4pt}
\renewcommand{\arraystretch}{0.96}
\caption{Selected paired visual-cue effects under the natural prompt. Net induced is cue-only minus plain-only false accepts over the same 300 invalid items; positive values increase false acceptance and negative values reduce it. Bold marks the largest shifts.}
\label{tab:cue_mcnemar_selected}
\rowcolors{2}{tablegray}{white}
\begin{tabular}{@{}llrr@{}}
\toprule
Model & Cue comparison & Net induced & McNemar $p$ $\downarrow$ \\
\midrule
GPT-4o & decorative + neutral & 13 & 0.00235 \\
Qwen3-VL-8B & Islamic authority & -34 & $5.65{\times}10^{-8}$ \\
Qwen3-VL-8B & neutral authority & -42 & $1.28{\times}10^{-9}$ \\
Qwen3-VL-32B & decorative & 45 & $6.27{\times}10^{-8}$ \\
Qwen3-VL-32B & Islamic authority & \textbf{-52} & $5.75{\times}10^{-11}$ \\
InternVL3.5-38B & Islamic authority & \textbf{62} & $8.47{\times}10^{-16}$ \\
InternVL3.5-38B & neutral authority & \textbf{62} & $2.26{\times}10^{-16}$ \\
\hiderowcolors
\bottomrule
\end{tabular}
\end{table}

\section{Paired prompt effects}

Table~\ref{tab:paired_prompt_effects} reports paired natural-versus-de-biased prompt effects on the same invalid items.

\begin{table}[H]
\centering
\small
\caption{Paired prompt effect on invalid-citation false accepts. Positive net rescue means the de-biased prompt corrected false accepts made under the natural prompt. Lower McNemar $p$ indicates stronger evidence of asymmetric paired flips.}
\label{tab:paired_prompt_effects}
\resizebox{\linewidth}{!}{
\rowcolors{2}{tablegray}{white}
\begin{tabular}{lrrrrr}
\toprule
Model & Invalid pairs & Natural-only false accept $\downarrow$ & De-biased-only false accept $\downarrow$ & Net rescue $\uparrow$ & McNemar $p$ $\downarrow$ \\
\midrule
GPT-4.1 & 1800 & 50 & 22 & 28 & 0.00129 \\
GPT-4o & 1800 & 57 & 32 & 25 & 0.01055 \\
GPT-5.5 & 1800 & \textbf{0} & \textbf{0} & 0 & 1.00000 \\
GPT-5.6 Terra & 1800 & 51 & 51 & 0 & 1.00000 \\
Gemma 4 12B-IT & 1800 & 36 & \textbf{0} & 36 & $2.91{\times}10^{-11}$ \\
InternVL3.5-38B & 1800 & 603 & \textbf{0} & \textbf{603} & $<10^{-10}$ \\
InternVL3.5-8B & 1800 & 73 & \textbf{0} & 73 & $<10^{-10}$ \\
MiniCPM-V 4.5 & 1800 & 50 & \textbf{0} & 50 & $<10^{-10}$ \\
Qwen2.5-VL-32B & 1800 & 17 & 3 & 14 & 0.00258 \\
Qwen2.5-VL-7B & 1800 & 61 & \textbf{0} & 61 & $<10^{-10}$ \\
Qwen3-VL-32B & 1800 & 164 & 2 & 162 & $<10^{-10}$ \\
Qwen3-VL-8B & 1800 & 50 & 2 & 48 & $6.12{\times}10^{-13}$ \\
\hiderowcolors
\bottomrule
\end{tabular}}
\end{table}

\paragraph{Reading the paired table.}
Natural-only false accepts are invalid images accepted under the natural prompt but corrected under the de-biased prompt. De-biased-only false accepts are regressions introduced by the de-biased prompt. Net rescue is the difference between these two counts, while lower McNemar's $p$ reports stronger evidence that the paired flips are asymmetric rather than random prompt noise.

\end{document}